\documentclass[sigconf]{acmartarxiv}

\AtBeginDocument{%
  }

\setcopyright{none}
\copyrightyear{2026}
\acmYear{2026}
\acmConference[MM '26]{Proceedings of the 34th ACM International Conference on Multimedia}{November 10--14, 2026}{Rio de Janeiro, Brazil}
\renewcommand\footnotetextcopyrightpermission[1]{}
\usepackage{amsmath}
\usepackage{booktabs}
\usepackage{multirow}
\usepackage{tabularx}
\usepackage[table]{xcolor}
\usepackage{algorithm}
\usepackage{algorithmic}
\microtypesetup{expansion=false}
\title{PCA: Persistence-Aware Compression and Aggregation for Fast Video Large Language Models}

\newcommand{\affone}{\textsuperscript{1}}
\newcommand{\affthree}{\textsuperscript{3}}
\newcommand{\affonetwoequal}{\textsuperscript{1,2,\textdagger}}
\newcommand{\affoneequal}{\textsuperscript{1,\textdagger}}
\newcommand{\affonefourfivecorr}{\textsuperscript{1,4,5,*}}
\author{%
  Zihan Song\affonetwoequal\quad
  Shuo Ye\affoneequal\quad
  Bo Zhao\affone\quad
  Ruixin Zhang\affthree\\
  Jiayu Zhang\affone\quad
  Shouhong Ding\affthree\quad
  Zitong Yu\affonefourfivecorr}
\affiliation[obeypunctuation=false]{%
  \institution{%
    \textsuperscript{1}Institute for Artificial Intelligence, Great Bay University\\
    \textsuperscript{2}Sun Yat-sen University\\
    \textsuperscript{3}Youtu Lab, Tencent\\
    \textsuperscript{4}Guangdong Provincial Key Laboratory of Intelligent Information Processing \& Shenzhen Key Laboratory of Media Security, Shenzhen University\\
    \textsuperscript{5}Dongguan Key Laboratory for Intelligence and Information Technology}
  \country{}
}

\renewcommand{\shortauthors}{Song et al.}

\begin{document}

\fancyhead[RO,LE]{}

\begin{abstract}
Despite advances in Video Large Language Models (VLLMs) that have displayed promising outcomes in video understanding, the redundancy in the long-duration frames remains a hindrance to efficient reasoning. This paper introduces a training-free \textbf{P}ersistence-Aware \textbf{C}ompression and \textbf{A}ggregation (PCA) method designed to preserve high-fidelity raw visual information before the encoding stage. PCA can be built on arbitrary VLLMs and consists of two modules: 1) A Dynamic Downsampling (DD) module that adaptively removes redundant frames by analyzing frame-wise similarity. 2) A Persistence-Aware Motion Enhancement (PAME) module that enriches each selected keyframe by aggregating the temporal context of its neighbors, ensuring that essential information is preserved even after aggressive frame reduction. Our approach substantially reduces the computation of long-context modeling, while enhancing the performance of the baseline model. Extensive experiments demonstrate that PCA consistently outperforms existing state-of-the-art approaches in both efficiency and accuracy, achieving a speedup of 1.8$\times$ to 2.5$\times$ compared to the baseline VLLM. The code is open-sourced at \href{https://github.com/Heisenberg10110/PCA}{\texttt{Heisenberg10110/PCA}}.
\end{abstract}



\maketitle

\begingroup
\renewcommand{\thefootnote}{}
\footnotetext{\textsuperscript{\textdagger} Zihan Song and Shuo Ye contributed equally.\newline\textsuperscript{*} Corresponding author: Zitong Yu.}
\endgroup

\section{Introduction}
Visual perception is an indispensable part of our daily lives. With the rapid advancement of large language models (LLMs) and multimodal learning techniques, machines are gaining a deeper understanding of the world around us~\citep{VLM_expanding,llama,gpt3,lion,zhao2025accelerating,ROD-MLLM,vcoder,seedbench,catplus,omniopsd,multimodal_deception_survey,emotion_gen_survey,phase_net,physllm,human_motion_survey,etag,preprompt}. In particular, Video LLMs (VLLMs) have demonstrated remarkable progress in understanding and reasoning complex video content~\citep{Video-LLaMA2,Video-ChatGPT,VLM_Llava-onevision,VLM_videochat,VLM_qwen2,add_vlm_1,add_vlm_Stream,add_vlm_agent,vlm_add_apollo,reflectr1}. 
However, when a video is decomposed into a sequence of frames and processed by large models, the computational complexity often grows quadratically with the number of frames, due to the attention mechanism in transformer-based architectures. To accelerate inference, some works~\citep{mobilevlm,mobilevlmv2} attempt to reduce model parameters to alleviate the rapidly increasing computational complexity. However, such approaches often lead to a notable drop in model performance. Therefore, improving inference efficiency without sacrificing model accuracy remains an important research direction for VLLMs.

\begin{figure}[t]
    \centering
    \includegraphics[width=\linewidth]{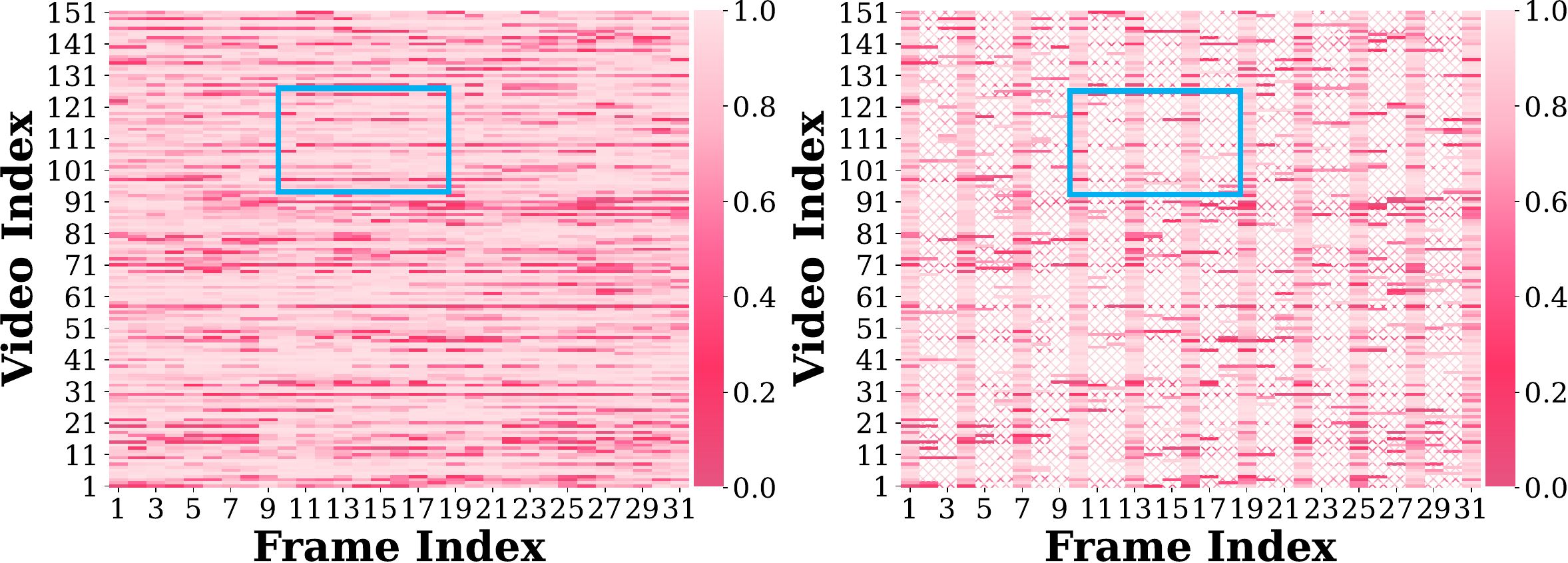}
    \vspace{-1.5em}
    \caption{\textbf{Pairwise similarity heatmap between adjacent video frames.} The left figure shows the pairwise similarity of adjacent frames in the original sequence. The right figure shows the result after applying our method, where most redundant frames are removed, high-discrepancy key frames are preserved, and discarded frames appear as masked positions.}
    \Description{A two-panel heatmap of adjacent-frame similarity. The left panel shows strong redundancy in the original video, while the right panel shows that the proposed method removes many redundant frames and keeps only key frames with larger visual changes.}
    \label{heatmap}
    \vspace{-0.9em}
\end{figure}

In prior studies~\citep{FASTV,LLaVA-PruMerge,DYCOKE,TempMe,TESTA,fan2026flashvid} on accelerating VLLMs, token pruning is a commonly adopted strategy, where the attention scores between visual tokens serve as the primary criterion for identifying and removing less informative tokens.
For example, FastV~\citep{FASTV} leverages the attention scores between predicted tokens and visual tokens during the prefilling stage to identify and prune redundant visual tokens, thereby accelerating the inference of large language vision models. In contrast, DyCoke~\citep{DYCOKE} observes that attention distributions in videos vary significantly over time and thus proposes a dynamic token pruning strategy that accounts for temporal changes. FlashVID~\citep{fan2026flashvid} further improves efficiency through training-free tree-based spatiotemporal token merging. 
However, the inherently sequential nature of videos leads to substantial temporal and spatial redundancy, a challenge that is particularly pronounced when encoding every frame in full resolution, resulting in excessive computational overhead. Moreover, most previous token pruning approaches overlook the redundancy between frames, focusing only on within-frame token selection, which leaves significant global redundancy untapped. 
Consequently, applying token pruning only after every frame has been fully encoded still incurs substantial overhead for VLLMs.

\begin{figure}[t]
    \centering
    \includegraphics[width=\linewidth]{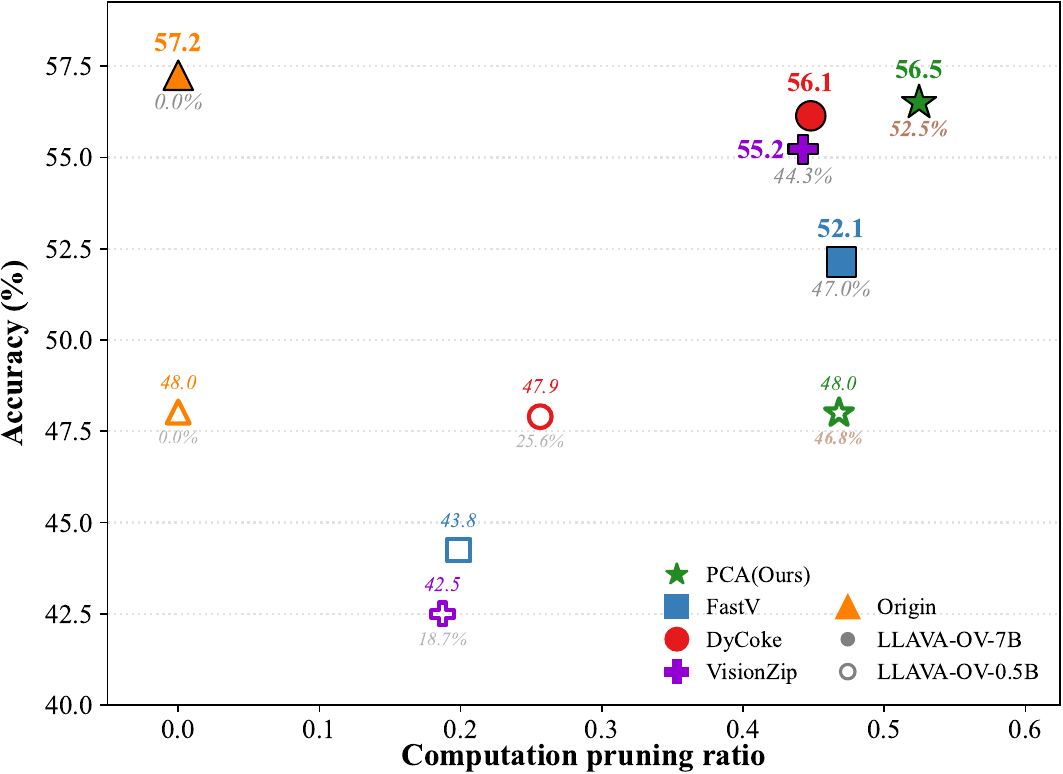}
    \vspace{-1.5em}
    \caption{\textbf{Comparison of models in terms of mean accuracy and total FLOPs on the MVBench dataset.}
Our proposed PCA substantially reduces computational cost while maintaining high model performance, outperforming DyCoke and FastV with fewer FLOPs. “Origin” denotes the LLaVA-OV baseline without pruning or compression. }
    \Description{A scatter plot comparing mean accuracy and total FLOPs on MVBench. PCA achieves a stronger accuracy-efficiency trade-off than the baseline compression methods, with lower FLOPs and competitive or better accuracy.}
    \label{fig:acc_vs_flops}
    \vspace{-0.8em}
\end{figure}

To tackle the challenge of global computational redundancy, we revisit the need to jointly reduce the overhead of both visual encoding and inference, treating the minimization of interframe redundancy before VLLM processing as a fundamental design principle. 
We propose the \textit{persistence-aware compression and aggregation method} (PCA).
1) \textbf{Dynamic Downsampling (DD)}: We perform dynamic frame downsampling by comparing inter-frame similarity, aiming to reduce the number of frames while retaining essential video content.
2) \textbf{Persistence-Aware Motion Enhencement (PAME)}: We introduce a lightweight module that enhances each selected frame in a persistence-aware manner, allowing it to carry dynamic cues from neighboring frames without any additional training.
Specifically, our method ensures that the model receives compact, yet semantically rich frames, compensating for the loss caused by frame reduction and boosting reasoning performance.
Based on this design, PCA enables video LLMs to reason over fewer but higher-quality frames, significantly reducing overall computation while maintaining strong performance.

Empirically, the proposed PCA achieves substantial acceleration in both the encoding and inference stages of video understanding and reasoning tasks, while maintaining strong performance. By aggressively reducing redundant frame input, PCA not only accelerates the LLM stage by 80\% to 110\%, but also speeds up the encoding process by a similar margin. As shown in Figure~\ref{heatmap}, we examine the distribution of pairwise similarity scores between adjacent frames, revealing that many consecutive frames are highly redundant in visual content. In Figure~\ref{fig:acc_vs_flops}, our method attains the highest accuracy among all approaches while operating at the lowest computational cost, with the computation reduced to only half of the original amount.\looseness=-1

Our contributions are summarized as follows:
\begin{itemize} 
    \item We propose a plug-and-play and training-free module that is capable of downsampling frames by intelligently leveraging temporal context information.
    \item We introduce an effective approach to mitigate the substantial information loss caused by aggressive frame reduction, ensuring that the visual representations retain rich temporal and contextual cues even under the low framerate settings.
    \item Experiments on multiple benchmarks show that our approach preserves high accuracy and achieves a substantial reduction in global computational overhead. Our method achieves superior robustness under sparse frame conditions, while maintaining accuracy with minimal degradation at low frame rates.
\end{itemize}

\section{Related Works}
\subsection{Video Large Language Models}

One common paradigm in Video large language models is to treat a video as a sequence of static frames~\citep{Video-LLaMA,Video-ChatGPT,VLM_videochat,InternVideo}. These models typically select a sparse subset of frames and extract visual features using pretrained vision encoders. The resulting features are then concatenated or averaged before being fed into an LLM. Representative examples include Video-LLaMA~\citep{Video-LLaMA}, Video-ChatGPT~\citep{Video-ChatGPT}, and InternVideo~\citep{InternVideo}. This approach leverages existing image-text alignment pretraining, allowing efficient reuse of vision-language models. However, by ignoring the temporal continuity of frames, such models may fail to capture motion patterns and dynamics critical to video understanding.
To better leverage temporal information, subsequent works incorporate time-aware mechanisms into the video-to-language pipeline. Some, like Video-LLaMA2~\citep{Video-LLaMA2}, TimeChat~\citep{TimeChat} and TEVL~\citep{tevl}, inject explicit temporal modules, such as temporal attention or recurrent blocks, either before or within the visual encoder. These components help the model understand the evolution of visual content across frames, thereby improving coherence and reasoning in temporally rich tasks.
However, although these methods achieve good performance, their utilization of video frames is not efficient enough.

\subsection{Efficient Multi-Modal Large Language Models}

Multi-modal large language models (MLLMs) have already demonstrated impressive capabilities in understanding and reasoning about complex and long-duration video content. However, this also results in substantial computational overhead during inference. In order to address this, some works adopt token pruning mechanisms to reduce the number of visual tokens passed to the language model. LLaVA-PruMerge~\citep{LLaVA-PruMerge} selects key visual tokens based on attention scores during image encoding, thereby reducing redundancy in the input. ToMe~\citep{ToMe} merges similar tokens to reduce the token count during inference. 
VisionZip~\citep{VisionZip} introduces a training-free strategy that prunes and aggregates tokens by leveraging the attention patterns within encoded video frame features, aiming to produce a compact and informative video representation.
FastV~\citep{FASTV} proposes an inference-time token selection method that allows the model to skip redundant visual features during inference.

Some studies introduce a broader range of strategies to address spatial and temporal redundancy. TempMe~\citep{TempMe} focuses on temporal redundancy by merging adjacent video clips. TESTA~\citep{TESTA} utilizes temporal and spatial aggregation to improve video language understanding while reducing computational redundancy.
DyCoke~\citep{DYCOKE} proposes a temporal-aware pruning method that captures the dynamic evolution of attention across frames. Despite these efforts, many existing methods rely on pre-encoded representations or static heuristics, limiting their ability to recover high-level semantics lost due to early pruning. Our work addresses this gap by explicitly enhancing retained frames through context-aware aggregation, thereby improving efficiency while preserving richer semantic content for downstream reasoning.

\section{Method}
\subsection{Background on Video Input for VLLMs}
Given a video, frames are usually sampled at a fixed frame rate (e.g. $f$ FPS), resulting in a sequence of frames $\mathcal{F} = [F_1, F_2, \dots, F_m]$. Each frame $F_i$ is individually encoded by a visual encoder into a set of embedding vectors $\mathbf{z_i} \in \mathbb{R}^{H_i \times d}$, where $H_i$ is the number of patch tokens for frame $i$, and $d$ is the embedding dimension. The sequence of visual representations is thus $\mathbf{Z}_v = [\mathbf{z}_1, \mathbf{z}_2, \ldots, \mathbf{z}_{m}]$. After being projected into the language embedding space, visual representations are fed into a pre-trained large language model.

In typical VLLMs, the majority of computation time is consumed by visual encoding and language inference.  
By downsampling the input frame rate to $\alpha f$ ($0 < \alpha < 1$), the number of visual tokens can be effectively reduced, thereby accelerating both stages.  
Although this may introduce some loss of visual information, it yields a substantial improvement in the response speed $\mathcal{V}$:
\begin{equation}
\Delta \mathcal{V}_{\text{rel}} = \frac{1 - \alpha}{\alpha} \times 100\%
\label{eq:delta_v_formula},
\end{equation}
where $\Delta \mathcal{V}_{\text{rel}}$ denotes the relative improvement, defined as:
\begin{equation}
\Delta \mathcal{V}_{\text{rel}} = \frac{\mathcal{V}' - \mathcal{V}}{\mathcal{V}}
\label{eq:delta_v_definition}.
\end{equation}
With $\mathcal{V}$ and $\mathcal{V}'$ representing the original and improved response speeds, respectively.

\begin{figure*}[t]
  \centering
  \includegraphics[width=\textwidth]{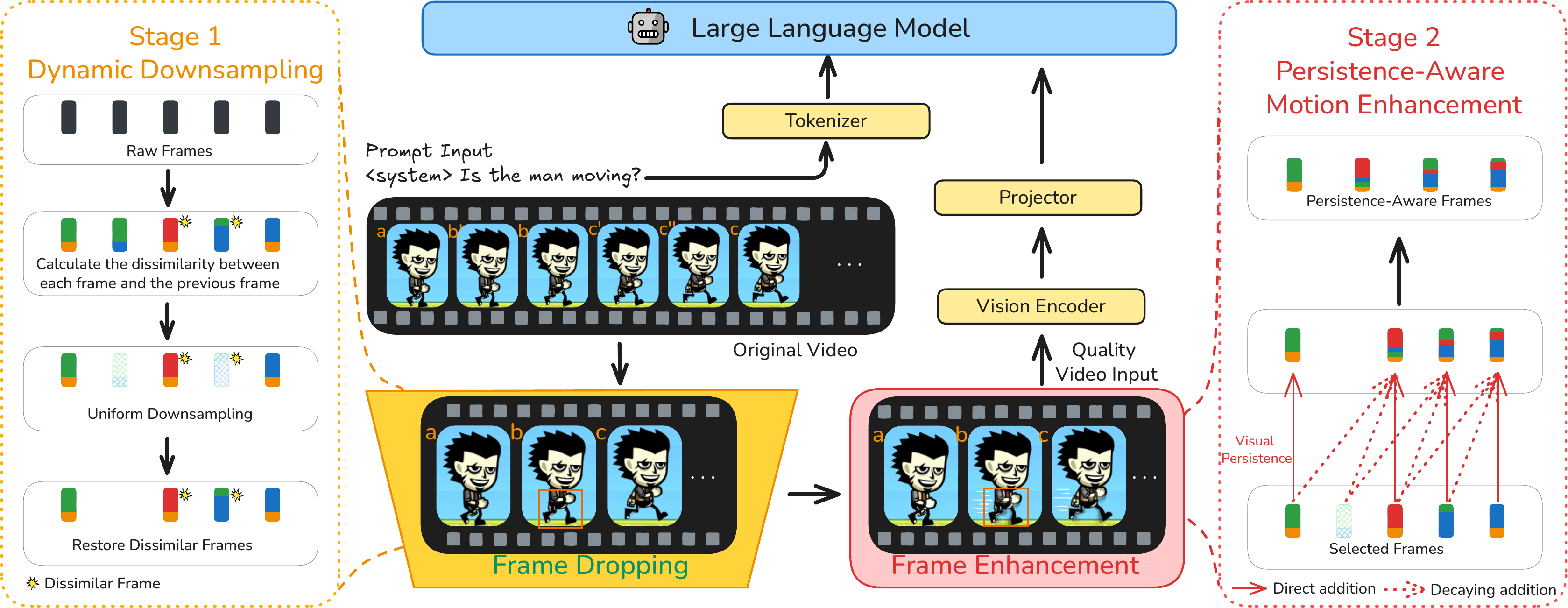}
  \caption{Overview of the proposed PCA framework, which consists of two main modules: Dynamic Downsampling and Persistence-Aware Motion Enhancement. Dynamic Downsampling sparsely selects key frames by measuring inter-frame dissimilarity, while Persistence-Aware Motion Enhancement improves each retained frame by aggregating information from its temporally adjacent neighbors. PCA is a plug-and-play, training-free approach to accelerate VLLMs. The different colors within the frames represent distinct pixel-level semantics.}
  \Description{An overview diagram of PCA with two stages. Dynamic Downsampling selects key frames based on inter-frame dissimilarity, and Persistence-Aware Motion Enhancement augments each retained frame with information from temporal neighbors before the frames are passed to the VLLM.}
  \label{framework}
  
\vspace{-0.8em}
\end{figure*}

\subsection{Framework Overview}
In the human visual system, the retina serves as the primary site of sensory preprocessing~\citep{retina}. Rather than functioning in a binary on-off manner, it displays a graded, decaying response: after each photon strike, the ensuing neural signals diminish over a brief temporal window~\citep{retina2}. This phenomenon, commonly referred to as \textit{visual persistence}~\citep{persistence}, not only smooths the transition between successive stimuli but also enhances the effective utilization of incoming visual information. Drawing inspiration from these neural mechanisms, we propose PCA, a training-free method to reduce redundant frames while preserving essential motion information. This method selectively extracts persistence-aware information, mirroring the retina's ability to efficiently process continuous visual input.

As illustrated in Figure~\ref{framework}, our approach employs a two-stage frame compression strategy. The first stage sparsely selects key frames. Based on these selected frames, the second stage enhances each frame by aggregating information from its temporally adjacent neighbors, forming persistence-aware representations. This design facilitates efficient extraction and selection of informative visual content prior to its integration into the VLLMs, thereby reducing redundancy and improving the quality of the resulting representations for downstream reasoning.

\subsection{Dynamic Downsampling}
In VLLMs, the video input contains significant temporal redundancy. Some activities persist across multiple frames with minimal variation, and backgrounds or static elements often remain nearly unchanged. We directly perform coarse-grained frame sampling by skipping every few frames. To compensate for potentially lost information, we supplement this sampling with the inclusion of key frames that exhibit substantial visual changes. This stage filters out consecutive redundant frames while preserving salient event transitions and visually distinctive moments throughout the video.

Given a video with $m$ frames $\mathcal{F} = [F_1, F_2, \ldots, F_m]$, we apply uniform downsampling by selecting one frame every $k$ frames. The resulting subset of frames:
\begin{equation}
\mathcal{F}' = [F_1, F_k, F_{2k}, \ldots, F_{lk}]
,\
\end{equation}
where $l = \left\lfloor \frac{m}{k} \right\rfloor$. 

To compensate for the potential loss of important visual information caused by downsampling, we construct a final selected frame set \( \mathcal{H} \), which combines the uniformly sampled frames \( \mathcal{F}' \) with an additional set \( \mathcal{G} \):
\begin{equation}
\mathcal{H} = \mathcal{F}' \cup \mathcal{G}
\label{eq:frame_union}.
\end{equation}

Here, \( \mathcal{G} \) consists of frames that exhibit substantial visual differences compared to their adjacent frames.

First, we define an ordered sequence:
\begin{equation}
\mathcal{P} = \left( (F_{i}, F_{i+1}, S(F_{i}, F_{i+1}) \;\middle|\; i = 1, 2, \ldots, m-1 \right),
\end{equation}
where the sequence is sorted by $S(F_{i}, F_{i+1})$ in ascending order. \( S(F_i, F_{i+1}) \) denotes the similarity between two consecutive frames \( F_i \) and \( F_{i+1} \):
\begin{equation}
S(F_i, F_{i+1}) = \frac{\mathbf{X}_i \cdot \mathbf{X}_{i+1}}{\|\mathbf{X}_i\| \, \|\mathbf{X}_{i+1}\|},
\label{eq:cosine_similarity}
\end{equation}
with $\mathbf{X_i}$ denoting the flattened pixel vector of frame \( F_i \).

Based on the restored rate $p$ ($0<p<1$), we select a subset $\mathcal{P'}$ from $\mathcal{P}$ as follows:
\begin{equation}
\mathcal{P}' = \left\{\, \pi_i \in \mathcal{P} \;\middle|\; \frac{i}{\left\| \mathcal{F} \setminus \mathcal{F}' \right\|} < p \,\right\},
\end{equation}
where $\left\| \mathcal{F} \setminus \mathcal{F}' \right\|$ represents the cardinality of the set difference between $\mathcal{F}$ and $\mathcal{F}'$. We finally obtain the restored frames:
\begin{equation}
\mathcal{G} = \left\{ F_i \;\middle|\; (F_{i-1}, F_{i}, S(F_{i-1}, F_{i})) \in \mathcal{P}' \right\}.
\end{equation}

This union strategy ensures that frames that capture meaningful visual transitions are retained, complementing the uniformly sampled set \( \mathcal{F}' \) and thereby providing more complete visual input for downstream processing.

\subsection{Persistence-Aware Motion Enhancement}
To further mitigate the representational loss caused by aggressive downsampling, we focus on enriching the semantic and temporal information embedded in each frame. Each frame captured by the camera primarily preserves the static appearance and spatial location of objects at a specific timestamp, but lacks temporal awareness of how these objects evolve across adjacent frames. This temporal myopia limits the model's ability to understand continuity, motion dynamics, and event evolution over time.

To address this, we introduce Persistence-Aware Motion Enhancement (PAME) as the second-stage solution. 
Designed as a plug-and-play, training-free module, PAME operates after the downstream frame selection step and enhances the temporal expressiveness of each retained frame.
By incorporating short-term memory from previous frames, PAME supplements missing motion cues and helps the model build a more coherent understanding of scene dynamics over time.

Assume that we have the original video frames denoted as \( \mathcal{F} = \{F_1, F_2, \dots, F_m\} \) and a sparse set of frames is selected as \( \mathcal{H} = \{F_{t_1}, F_{t_2}, \dots, F_{t_n}\} \subset \mathcal{F} \) for downstream processing.

\begin{algorithm}[t]
\caption{PCA: Persistence-Aware Compression and Aggregation}
\label{alg:PCA}
\begin{algorithmic}
\STATE \textbf{Input:} frame sequence \texttt{frames}, downsampling step \texttt{step}, restore rate \texttt{restore\_rate}, window size \texttt{window}, decay factor \texttt{decay}
\STATE \textbf{Output:} frame sequence \texttt{enhanced\_frames}
\STATE $\mathcal{F}' \leftarrow$ \texttt{frames[::step]}\hfill \textcolor{gray}{\# uniform downsampling}
\STATE $\mathcal{P}  \leftarrow$ frames that are not selected in $\mathcal{F}'$
\STATE For each frame $p_i$ in $\mathcal{P}$, compute cosine similarity with the previous frame in \texttt{frames}, store as \texttt{similarity}
\STATE $r \leftarrow \mathrm{int}(\texttt{restore\_rate} \times |\mathcal{P}|)$
\STATE Obtain indices of the $r$ frames in $\mathcal{P}$ with the lowest similarity (ascending order), store as \texttt{idx}
\STATE $\mathcal{G} \leftarrow$ select frames in $\mathcal{P}$ according to \texttt{idx}
\STATE Merge $\mathcal{F}'$ and $\mathcal{G}$, then sort by the original frame order to obtain $\mathcal{H}$
\FOR{each frame $h$ in $\mathcal{H}$}
    \STATE \texttt{enhancement} $\leftarrow 0$
    \STATE \texttt{start\_idx} $\leftarrow \max(0, h.\mathrm{idx} - \texttt{window})$
    \FOR{$i = \texttt{start\_idx}$ \textbf{to} $h.\mathrm{idx}$}
        \STATE \texttt{weight} $\leftarrow$ \texttt{decay}$^{(h.\mathrm{idx}-i)}$
        \STATE \texttt{enhancement} += \texttt{weight} $\times$ \texttt{frames}[i]
        \STATE \texttt{weight\_sum} += \texttt{weight}
    \ENDFOR
    \STATE $h \leftarrow \frac{\texttt{enhancement}}{\texttt{weight\_sum}}$
\ENDFOR
\STATE \textbf{return} $\mathcal{H}$
\end{algorithmic}
\end{algorithm}

For each selected frame \( F_i \in \mathcal{F} \), we denote its representation at the pixel level as \( \mathbf{Y_i} \in \mathbb{R}^{H \times W \times C} \), where \( H \), \( W \), and \( C \) indicate the height, width, and number of channels, respectively.
we generate an enhanced representation $\mathbf{Y_i'}$ via our persistence-aware method as:
\begin{equation}
\mathbf{Y_i'} = \mathrm{PAME}(F_i, \mathcal{F}).
\end{equation}

We instantiate this enhancement as a weighted accumulation over preceding frames:
\begin{equation}
\mathrm{PAME}(F_{t_k}, \mathcal{F}) =
\frac{\sum_{i=1}^{t_k} \ell_i \cdot \mathbf{Y_i}}
     {\sum_{i=1}^{t_k} \ell_i}
\label{eq:PAME}
,
\end{equation}
where the weight \( \ell_i \) is defined as:
\begin{equation}
\ell_i =
\begin{cases}
\alpha^{t_k - i}, & \text{if } t_k - i < j \\
0,                & \text{otherwise}.
\end{cases}
\end{equation}
Here, \( \alpha \) is a fixed decay coefficient, and \( j \) denotes the size of the temporal window used for temporal aggregation.


We apply the operation defined in Eq.~(\ref{eq:PAME}) to each frame \( F_i \in \mathcal{H} \), resulting in the final set \( \mathcal{T} \) of frames fed into the VLLM. The pixel-level representations of \( \mathcal{T} \), denoted as \( \mathbf{Y}(\mathcal{T}) \), are given by the formulation below:
\begin{equation}
\mathbf{Y}(\mathcal{T}) = \left\{ \mathbf{T}_i \,\middle|\, \mathbf{T}_i = \mathrm{PAME}\left (F_i, \mathcal{F}\right),\, F_i \in \mathcal{H} \right\}.
\label{eq:PAME_outputs}
\end{equation}

Through our persistence-aware frame selection and enhancement strategy, the input sequence to VLLMs becomes both computationally efficient and semantically expressive.

\begin{table*}[t]
\centering
\small
\vspace{-0.8em}
\caption{\textbf{Performance of PCA on LLAVA-OV.} $K$, $J$ and $P$ are three hyper-parameters of our method:
$K$ denotes the interval for uniform downsampling (i.e. one frame is sampled for every $K$ frames);
$J$ represents the size of the temporal window for the weighting function;
$P$ denotes the rate of frames with the highest dissimilarity that are additionally selected. B, M and R represent BLEU, METEOR, and ROUGE-L scores. For all the values, except FLOPS and ratio, the higher is better. FLOPs are computed as the total sum, accounting for the vision encoder, LLM inference, and associated modules. The best result among token pruning methods for each metric is shown in \textbf{bold}, and the second best is \underline{underlined}.}
\vspace{-0.8em}
\begin{tabular}{l|cc|c c c c c c c |cc}
\toprule[1.3pt]
\multicolumn{1}{c|}{Method} & \multicolumn{2}{c|}{Efficiency} & MVBench & PercepTest & \multicolumn{3}{c}{YouCook2} & \multicolumn{2}{c|}{VideoMME} & \multicolumn{2}{c}{Avg.} \\
~ &FLOPs & Ratio & avg. & val & B & M & R & wo & w-subs & Score & \% \\
\midrule
LLaVA-OV-0.5B  & 16.5 & 100\% & 48.03 & 49.03 & 4.27 & 9.03 & 14.81 &   43.96 & 43.30 & 30.35 & 100\\
+ FastV \textit{[ECCV 2024]}  & 13.3 & 80.1\% & 43.75 & 43.07 & 3.73 & 5.24 & 11.06  & 40.48 & 39.22 & 26.65 & 87.8\\
+ VisionZip \textit{[CVPR 2025]}   & 13.4 & 81.3\% & 42.50 & 43.83 & \textbf{4.83} & 7.40 & 13.47   & 38.44 & 37.70 &26.88 & 88.6  \\
+ DyCoke \textit{[CVPR 2025]}   & 12.3  & 74.3\% & \underline{47.92} & \underline{49.18} & 3.86 & 6.84 & 12.57  & \underline{43.56} & 42.82 & 29.54 & 97.3\\
+ FlashVID \textit{[ICLR 2026]}   & 11.2  & 72.5\% & 46.10 & 45.62 & 4.22 & 7.96 & 11.52  & 43.21 & 41.35 & 28.56 & 94.1\\
\rowcolor{blue!5}+ \textbf{PCA} ($K$=2,$J$=3,$P$=0.2)  & 8.8 & 53.4\% & \textbf{48.03} & \textbf{49.68} & \underline{4.33} & \textbf{8.31} & \textbf{14.16}  & 43.48 & \underline{42.85} & \textbf{30.12} & \textbf{99.3}\\
\rowcolor{blue!5}+ \textbf{PCA} ($K$=3,$J$=3,$P$=0.4)   & 6.8 & 40.9\% & 47.14 & 48.84 & 4.27 & \underline{8.02} & \underline{13.77}  & \textbf{43.93} & \textbf{43.04} &\underline{29.86} & \underline{98.4}\\
\midrule
LLaVA-OV-7B   & 100.2 & 100\% & 57.25 & 57.10 & 5.40 & 9.44 & 16.05   & 58.48 & 57.89 & 37.37 & 100\\
+ FastV \textit{[ECCV 2024]}   & 53.1  & 53.0\% & 52.14 & 51.75 & 3.74 & 5.00 & 10.44   & 51.25 & 50.96 & 32.18 & 86.1\\
+ VisionZip \textit{[CVPR 2025]}  & 55.8  & 55.7\% & 55.22 & 55.89 & 4.87 & 6.98 & 13.10  & 54.96 & 53.63 & 34.95 & 93.5\\
+ DyCoke \textit{[CVPR 2025]}  & 42.6 & 42.5\% & 56.14 & 56.99 & 5.39 & 9.44 & 16.04   & \underline{56.37} & \underline{55.96} & 36.62 & 98.0\\
+ FlashVID \textit{[ICLR 2026]}   & 45.3  & 45.2\% & 53.65 & 56.55 & 5.38 & 9.81 & 14.63  & 55.68 & 52.51 & 35.46 & 94.5\\
\rowcolor{blue!5}+ \textbf{PCA} ($K$=2,$J$=3,$P$=0.2) & 53.8 & 53.7\%& \textbf{56.69} & \textbf{57.08} & \underline{5.62} & \underline{9.79} & \underline{16.48}  & \textbf{57.00} & \textbf{56.04} & \textbf{36.96} & \textbf{98.9} \\
\rowcolor{blue!5}+ \textbf{PCA} ($K$=3,$J$=3,$P$=0.4) & 41.4 & 41.4\% & \underline{56.33} & \underline{57.00} & \textbf{5.65} &\textbf{ 9.83} & \textbf{16.52}   & 56.11 & 55.70  & \underline{36.73} & \underline{98.3}\\

\bottomrule[1.3pt]
\end{tabular} 
\vspace{-0.8em}
\label{performance_1} 
\end{table*}

\section{Experiments}

\subsection{Datasets}
We validate our proposed framework on four video understanding benchmarks that cover multimodal reasoning, temporal reasoning, video captioning, and real-world scenarios.
\textbf{VideoMME}~\citep{videomme} evaluates the comprehensive multimodal reasoning ability across a diverse range of question types, testing the ability of the model to integrate and understand visual and textual information.
\textbf{PerceptionTest}~\citep{perceptiontest} is designed to systematically evaluate fine-grained perceptual understanding of video content. Focuses on the model's ability to care for and reason about subtle details, minor variations, and nuanced visual signals that can easily be overlooked. The benchmark comprises a diverse set of tasks that require precise discrimination of visual elements.
\textbf{MVBench}~\citep{mvbench} is a benchmark designed to evaluate the multiview reasoning capability of a model. 
Rather than focusing solely on frame-level understanding, MVBench tests whether a model can construct a coherent representation from heterogeneous inputs.
\textbf{YouCook2}~\citep{youcook2} comprises instructional cooking videos paired with natural language captions, serving as a benchmark for evaluating real-world understanding and multimodal alignment in instructional scenarios.

\subsection{Baselines}
We compare our method with three recent training-free visual token compression approaches. 
(1) FastV~\citep{FASTV} accelerates inference by deleting redundant visual tokens at inference time, where the importance of each token is determined by its attention score with respect to the predicted tokens during the prefilling stage. 
(2) VisionZip~\citep{VisionZip} selects informative tokens based on attention distribution and merges the remaining tokens according to spatial or semantic similarity, thus reducing the redundancy of visual tokens.
(3) DyCoke~\citep{DYCOKE} performs temporal-aware token pruning for video understanding. By tracking the evolution of attention scores across frames, DyCoke removes redundant visual tokens while preserving those that introduce new semantic information.

\subsection{Implementation Details}
We implement the proposed PCA on LLaVA-OneVision-0.5B and 7B models~\citep{VLM_Llava-onevision}. We also conduct experiments on VideoChatGPT~\citep{Video-ChatGPT} and InternVL2~\citep{internVL2} to demonstrate the plug-and-play generalizability of our method across distinct VLM architectures. 
To ensure a fair comparison, the pruning ratios are determined based on the total FLOPs calculated from both the encoder and the LLM components. For video input, we follow the official LLaVA-OneVision setting with 32 original frames and set $N_v = 196$.
Unless otherwise specified, all compared methods use the same video length and backbone configuration.

\subsection{Results}
The experimental results are shown in Table~\ref{performance_1}. Our method outperforms existing methods such as FastV, DyCoke and VisionZip, while achieving lower overall computational cost. This advantage is especially evident for the 0.5B model, where the encoding stage accounts for the majority of total computation. Other methods focus on token pruning in the LLM stage. Although this reduces the computation for the LLM, the overall efficiency remains limited by the heavy computational burden of the encoder. Furthermore, excessive token pruning in later stages often leads to a noticeable decrease in model performance. In contrast, by applying frame pruning before encoding, our approach greatly reduces the total amount of computation. On the long-video benchmark MVBench, PCA improves the accuracy by approximately 2\% over state-of-the-art methods while achieving higher efficiency. Notably, PCA achieves a reduction in FLOPs of 40\% to 50\% with negligible loss of precision.  These results highlight the advantage of performing compression before encoding, demonstrating the practical benefits of PCA in enabling video large language models to be more efficient.
This trend further indicates that moving compression before visual encoding is especially important when the encoder dominates the end-to-end cost, because later token pruning alone cannot remove that bottleneck.


\subsection{Ablation Study}

\noindent \textbf{Effectiveness of Dynamic Downsampling.}
We conduct extensive ablation experiments to evaluate the effectiveness of Dynamic Downsampling and PAME.
As shown in Table~\ref{ablation}, Dynamic Downsampling selects informative frames based on inter-frame similarity, allowing the model to discard redundant frames while preserving critical temporal cues.
Compared with random pruning, our method yields better performance on video understanding benchmarks, indicating that the improvement comes from informed frame selection rather than merely reducing the number of input frames.

\begin{table*}[t]
\centering
\small
\vspace{-0.8em}
\caption{
\textbf{Ablation study of our PCA method.} Further ablation results can be found in the appendix.
}
\vspace{-0.8em}
\begin{tabular*}{\textwidth}{@{\extracolsep{\fill}}l|cccc|c c c c c c cc}
\toprule
\multicolumn{1}{c|}{Method} & \multicolumn{4}{c|}{ Settings and Efficiency} & MVbench & PercepTest & \multicolumn{3}{c}{YouCook2} &  \multicolumn{2}{c}{VideoMME} \\
~ & $K$ & $J$ & $P$ & FLOPs & avg. & val & B & M & R & wo & w-subs \\
\midrule
LLaVA-OV-0.5B & - & - & - & 16.5 &  48.03 & 49.03 & 4.27 & 9.03 & 14.81 &   43.96 & 43.30  \\
Random Pruning & - & - & - & 8.3 & 46.97 & 48.62 & 4.20 &8.12&13.83&43.74&42.03 \\
\textbf{PCA (Ours)} & 2 & 2 & 0.4 & 9.3 & \textbf{48.06} & 49.70 & \textbf{4.33} & \textbf{8.31} & \textbf{14.15} & 43.63 & 42.48 \\
\textbf{PCA (Ours)} & 2 & 4 & 0.4 & 9.3 & 48.00 & \textbf{49.66} & 4.32 & 8.30 & 14.13 & 43.59 & 42.48 \\
\textbf{PCA (Ours)} & 3 & 3 & 0.4 & 6.8 & 47.14 & 48.84 & 4.27 & 8.02 & 13.76 & 43.93 & 43.04 \\
\textbf{PCA (Ours)} & 3 & 3 & 0.6 & 7.8 & 47.06 & 48.80 & 4.27 & 8.02 & 13.79 & \textbf{44.30} & \textbf{43.22} \\
\textbf{PCA (Ours)} & 4 & 3 & 0.4 & 5.7 & 45.78 & 48.11 & 4.19 & 7.71 & 13.29 & 43.00 & 41.44 \\
\midrule
LLaVA-OV-7B & - & - & - & 100.2 & 57.25 & 57.10 & 5.40 & 9.44 & 16.05   & 58.48 & 57.89  \\
Random Pruning & - & - & - & 50.1 & 55.75 & 56.37 & 5.45 &9.61 & 16.25 &54.51&54.62 \\
\textbf{PCA (Ours)} & 2 & 2 & 0.4 & 56.9 & 56.56 & 57.05 & 5.64 & 9.80 & 16.51 & 56.63 & 56.15 \\
\textbf{PCA (Ours)} & 2 & 4 & 0.4 & 56.9 & \textbf{56.64} & \textbf{57.05} & 5.65 & 9.82 & 16.53 & 56.63 & 56.15 \\
\textbf{PCA (Ours)} & 3 & 3 & 0.4 & 41.4 & 56.33 & 57.00 & 5.65 & 9.82 & 16.52 & 56.11 & 55.70 \\
\textbf{PCA (Ours)} & 3 & 3 & 0.6 & 47.6 & 56.39 & 57.04 & \textbf{5.66} & \textbf{9.85} & \textbf{16.53} & \textbf{56.85} & \textbf{56.30} \\
\textbf{PCA (Ours)} & 4 & 3 & 0.4 & 35.3 & 55.36 & 56.10 & 5.41 & 8.88 & 15.26 & 54.59 & 54.56 \\
\bottomrule

\end{tabular*}
\vspace{-0.8em}
\label{ablation}

\end{table*}

\begin{figure}[!t]
    \vspace{-0.8em}
    \centering
    \includegraphics[width=1\linewidth]{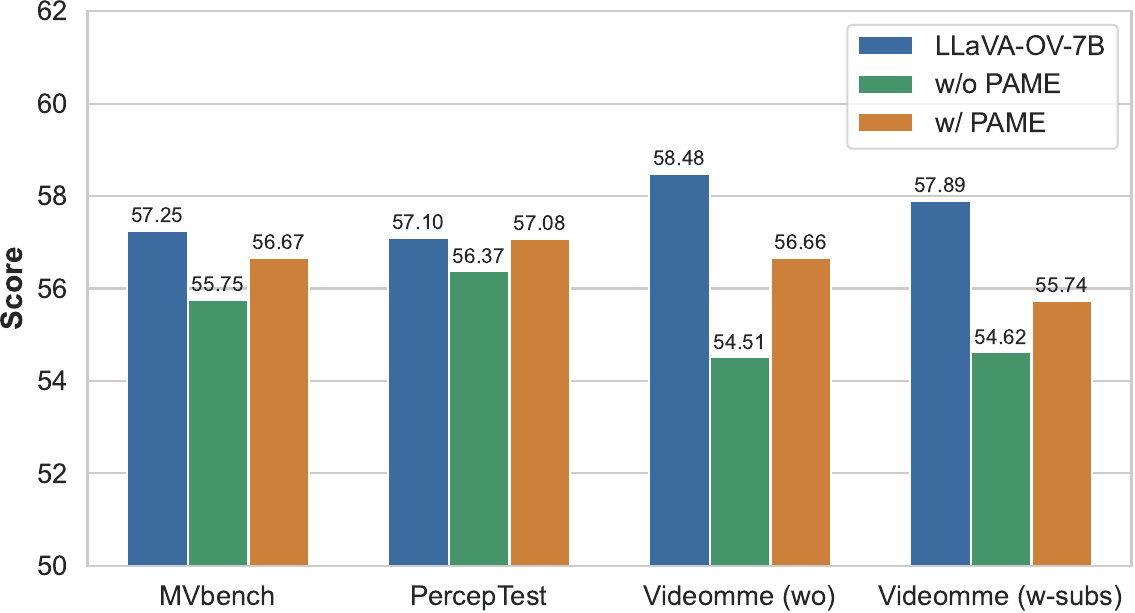} 
\caption{\textbf{Ablation study on the effect of the PAME.} 
The results are evaluated on LLaVA-OV-7B at a 50\% frame rate across video understanding benchmarks.
}
    \Description{A plot showing the effect of adding or removing the PAME module under reduced frame rate. The version with PAME performs better across the evaluated video understanding benchmarks.}
    \label{woPAME}
    \vspace{-0.8em}
\end{figure}
\vspace{0.3em}


 \vspace{0.2em}
\noindent \textbf{Effectiveness of the PAME Module.} In addition, we analyze the impact of the PAME module on the overall performance of our model. In Table~\ref{ablation}, increasing the number of aggregated frames leads to a noticeable performance improvement in the 7B model, while the effect is less significant in the 0.5B model. This suggests that models with larger parameter scales can better understand persistence-enhanced frames aggregated from more temporal neighbors, as they possess greater capacity to model subtle dependencies. As shown in Figure~\ref{woPAME}, at low frame rates, performance drops notably, indicating reduced temporal reasoning ability. The improvement observed on temporally focused benchmarks such as MVBench further indicates that PAME enhances the model's capacity for temporal reasoning. This result shows that our approach effectively makes frames persistence-aware, improving the model's ability to exploit temporal information and preserve essential features across neighboring frames.

\begin{table}[t]
\centering
\small
\caption{\textbf{Sensitivity analysis of the decay coefficient.}Other hyperparameters follow $K=2$, $J=3$, and $P=0.2$. Experiments are conducted on the LLaVA-OV-7B model.}
 
\setlength{\tabcolsep}{11.1pt} 

\begin{tabular}{l|c|c|cc}
\toprule
Method & $\alpha$ & MVBench & \multicolumn{2}{c}{VideoMME} \\
 &  & avg. & wo & w-subs \\
\midrule
\textbf{PCA} & 0.0   & 55.41 & 56.33 & 55.75 \\
\textbf{PCA} & 0.1 & 56.69 & 57.00 & 56.04 \\
\textbf{PCA} & 0.2 & 56.55 & 56.50 & 55.82 \\
\textbf{PCA} & 0.3 & 55.45 & 55.96 & 55.70 \\
\textbf{PCA} & 1.0 & 44.68 & 47.44 & 47.96 \\
\bottomrule
\end{tabular}

\label{ablation-alpha}
\vspace{-0.8em}
\end{table}

\noindent \textbf{Sensitivity analysis of the decay coefficient.} As shown in Table~\ref{ablation-alpha}, incorporating our persistence mechanism enhances the model's understanding ability. The performance remains largely stable across different values of the decay coefficient, indicating that the method is robust to variations in this hyperparameter. Notably, when $\alpha = 1$, the accuracy drops substantially, suggesting that a naïve uniform averaging introduces noise and dilutes informative temporal cues, further highlighting the effectiveness of our exponentially decayed weighting scheme for temporal aggregation.
This pattern suggests that moderate decay values preserve useful short-range motion cues while preventing stale information from distant frames from overwhelming the retained keyframe.

\begin{table}[htb]
\centering
\caption{\textbf{Performance comparison on different VLMs.} Results are evaluated on the MVBench benchmark.}
\footnotesize
\setlength{\tabcolsep}{3pt}
\resizebox{\columnwidth}{!}{
\begin{tabular}{l|cccccc}
\toprule
VLM & Origin & PCA & DyCoke & FastV & FlashVID & VisionZip \\
\midrule
InternVL2 & 40.27 & 39.07 & 35.75 & 38.63 & 36.71 & 37.96 \\
VideoChatGPT & 58.60 & 57.22 & 49.51 & 53.47 & 52.62 & 55.45 \\
\bottomrule
\end{tabular}
}
\label{vlm-compare}
\end{table}

\subsection{Generalization to Other VLM Architectures}
To verify the plug-and-play nature of PCA beyond the LLaVA-OneVision family, we further evaluate it on two additional VLM architectures, VideoChatGPT and InternVL2. 
To assess model-agnostic applicability, we report results on the benchmark MVBench, where all models are uniformly pruned by 30\%. 
As shown in Table~\ref{vlm-compare}, the results demonstrate that PCA outperforms the baselines on both VLMs, demonstrating strong generalizability to heterogeneous VLM designs.
This result suggests that the benefit of PCA mainly comes from input-side compression and temporal enhancement, rather than from assumptions tied to a specific VLM architecture.

\section{Model Analysis}
\begin{figure}[htb]
  \centering
  \includegraphics[width=1\linewidth]{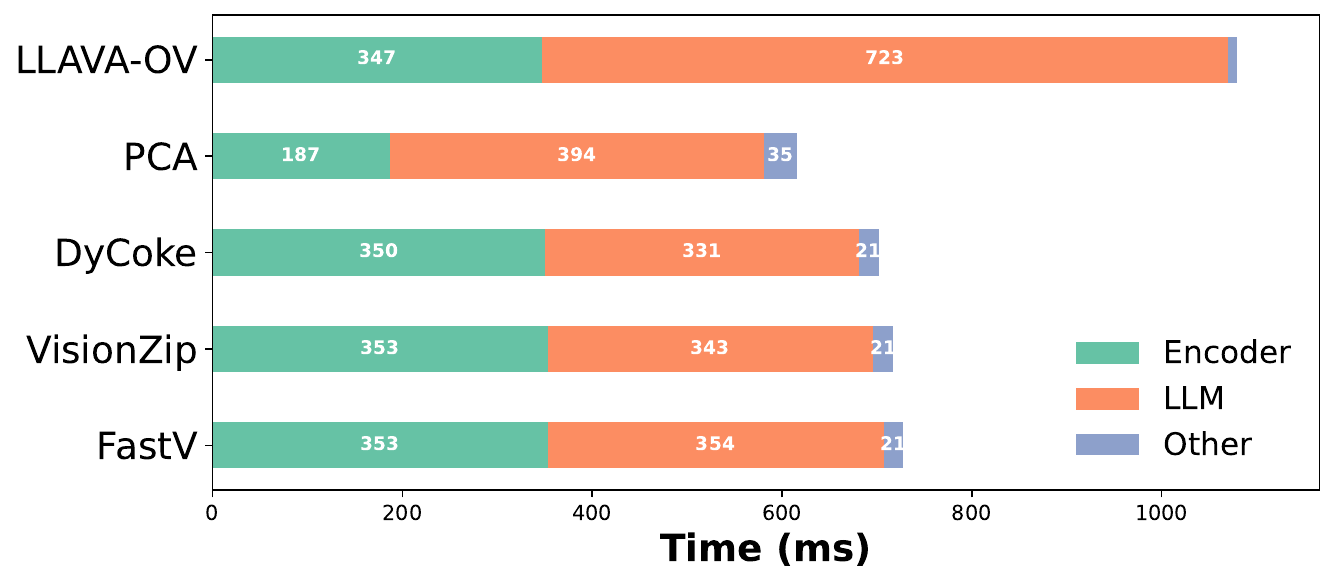}
  \vspace{-1.5em}
  \caption{
    \textbf{Comparison of component-wise time consumption across methods.} 
    The chart illustrates the time distribution for encoder, LLM, and other components in different visual language models. Our method achieves a significant reduction in both encoder and LLM computation, highlighting the advantage of pre-encoding pruning.
}
  \Description{A component-wise timing comparison across methods, separating encoder time, LLM time, and other overhead. PCA reduces both encoder and LLM computation and achieves the best overall efficiency.}
  \label{TIME}
  \vspace{-0.8em}
\end{figure}

\noindent \textbf{Factors Influencing Model Efficiency.} We investigated various factors that affect the overall efficiency of the proposed model. As illustrated in the Figure~\ref{TIME}, the majority of the time spent by Visual Large Language Models (VLLMs) on video understanding and reasoning tasks is concentrated in the encoder and LLM inference stages. Conventional pruning methods are highly effective in accelerating the LLM inference stage by reducing the number of visual tokens processed by the language model. However, because these pruning operations typically occur after the encoding stage, they do not reduce the computation time required by the encoder, which remains a significant bottleneck for overall efficiency. In contrast, our approach performs pruning prior to the encoding stage, selectively retaining only the most relevant frames or tokens from the video input. This early stage pruning substantially reduces the input size for both the encoder and the LLM, thereby achieving efficiency gains in both stages. 
Although our method does not achieve the fastest inference time at the LLM stage alone compared to some token-pruning baselines, the substantial reduction in encoder computation leads to an overall speedup of approximately 2$\times$. This makes PCA the fastest model among all compared methods.
As a result, our method delivers end-to-end improvements in computational efficiency without sacrificing accuracy. 
Furthermore, by alleviating computational load on the encoder, our approach is better suited for large-scale or resource-constrained video understanding tasks.
The computational overhead introduced by PCA grows only linearly with the number of frame patches, while the overall pipeline is dominated by the vision encoder and the LLM, whose computational cost increases superlinearly. As a result, the relative cost of PCA becomes progressively negligible as video length or resolution increases.

 \vspace{0.3em}

\noindent \textbf{Performance under Sparse Frame Conditions.} To assess the robustness of our approach under challenging conditions, we conduct experiments in which only a fraction of the original video frames are retained as input. 
In particular, we provide the VLLM with only a certain framerate ratio of the original frames, thus reducing temporal density and simulating severe frame sparsity.
\begin{figure}[t]
\vspace{-0.8em}
    \centering
    \includegraphics[width=1\linewidth]{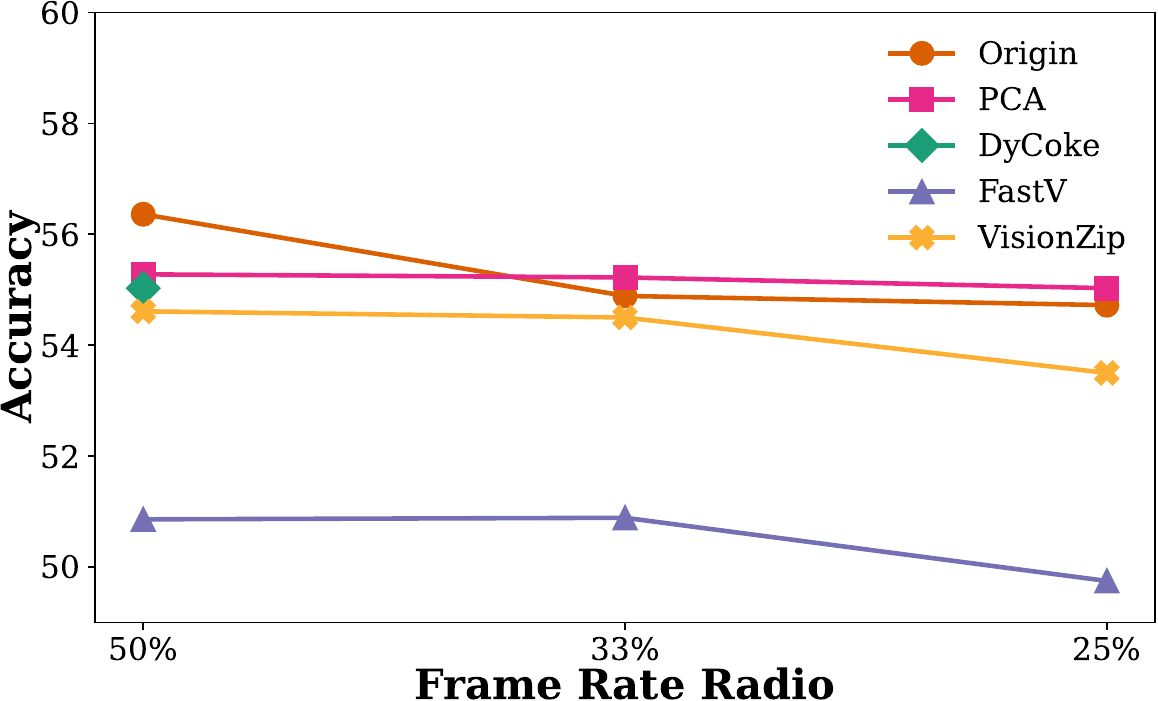} 
\caption{\textbf{Comparison of Methods on Sparse Frame Conditions.}
Results are evaluated on MVBench, with LLaVA-OV-7B as the baseline model. 
The frame rate ratio indicates the proportion of frames retained as input to the LLM, reflecting model robustness under limited frame availability.}
    \Description{A line chart comparing model accuracy under different retained frame-rate ratios on MVBench. PCA maintains the highest accuracy as fewer frames are kept, showing stronger robustness to sparse-frame inputs than the baselines.}
    \label{sparse-frame-performance}
    \vspace{-1.5em}
\end{figure}
\begin{figure}[b]
\vspace{-0.5em}
    \centering
    \includegraphics[width=\linewidth]{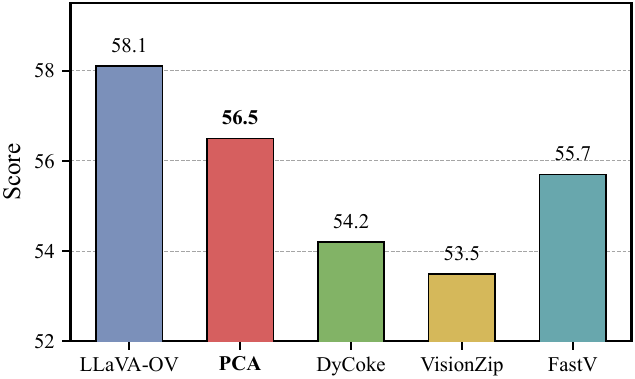}
    \vspace{-1.0em}
    \caption{\textbf{Performance on the top 20\% longest videos.}
    Results are evaluated on the longest 20\% videos in MVBench, with LLaVA-OV-7B as the backbone.}
    \Description{A bar chart comparing methods on the longest 20 percent of videos in MVBench. PCA remains close to the full LLaVA-OV baseline and outperforms several compression baselines on long-video evaluation.}
    \vspace{-0.8em}
    \label{long-video-performance}
\end{figure}
This setting directly compares model resilience under highly discontinuous visual inputs, a challenging case for temporal modeling.
As illustrated in Figure~\ref{sparse-frame-performance}, PCA maintains high accuracy even when the input becomes sparse, whereas the baselines degrade more noticeably as fewer frames are retained.
This gap reveals a limitation of prior approaches: they struggle to integrate temporally fragmented information and cannot maintain stable performance when the inputs are noncontiguous.
In contrast, PCA infers and reconstructs missing temporal context, producing a more coherent internal representation of the video.
This persistence-aware enhancement preserves essential information and is consistent with human visual perception, where the brain can interpolate missing details and still perceive temporal continuity under sparse inputs.
This remarkable result demonstrates that PCA's persistence-aware aggregation mechanism can effectively compensate for substantial information loss, highlighting its robustness in real-world scenarios characterized by incomplete frame sequences.


\noindent \textbf{Performance on Long Videos.}
To test whether repeated persistence-aware aggregation accumulates errors on long inputs, we isolate the top 20\% longest videos in MVBench and reevaluate all methods under the same protocol.
As shown in Figure~\ref{long-video-performance}, PCA achieves 56.5 accuracy, remaining within 1.6 points of the full LLaVA-OV baseline (58.1).
It also outperforms DyCoke (54.2) and VisionZip (53.5), while remaining competitive with FastV (55.7).
These results suggest that long videos may amplify accumulation effects in principle, but the actual degradation of PCA is limited.
The method remains competitive even under extended temporal contexts.
This behavior is consistent with the local design of PAME: each retained frame aggregates only short-range temporal evidence, which strengthens motion cues without allowing errors to accumulate unboundedly over long horizons.

\begin{figure}[t]
\vspace{-0.5em}
    \centering
    \includegraphics[width=\linewidth]{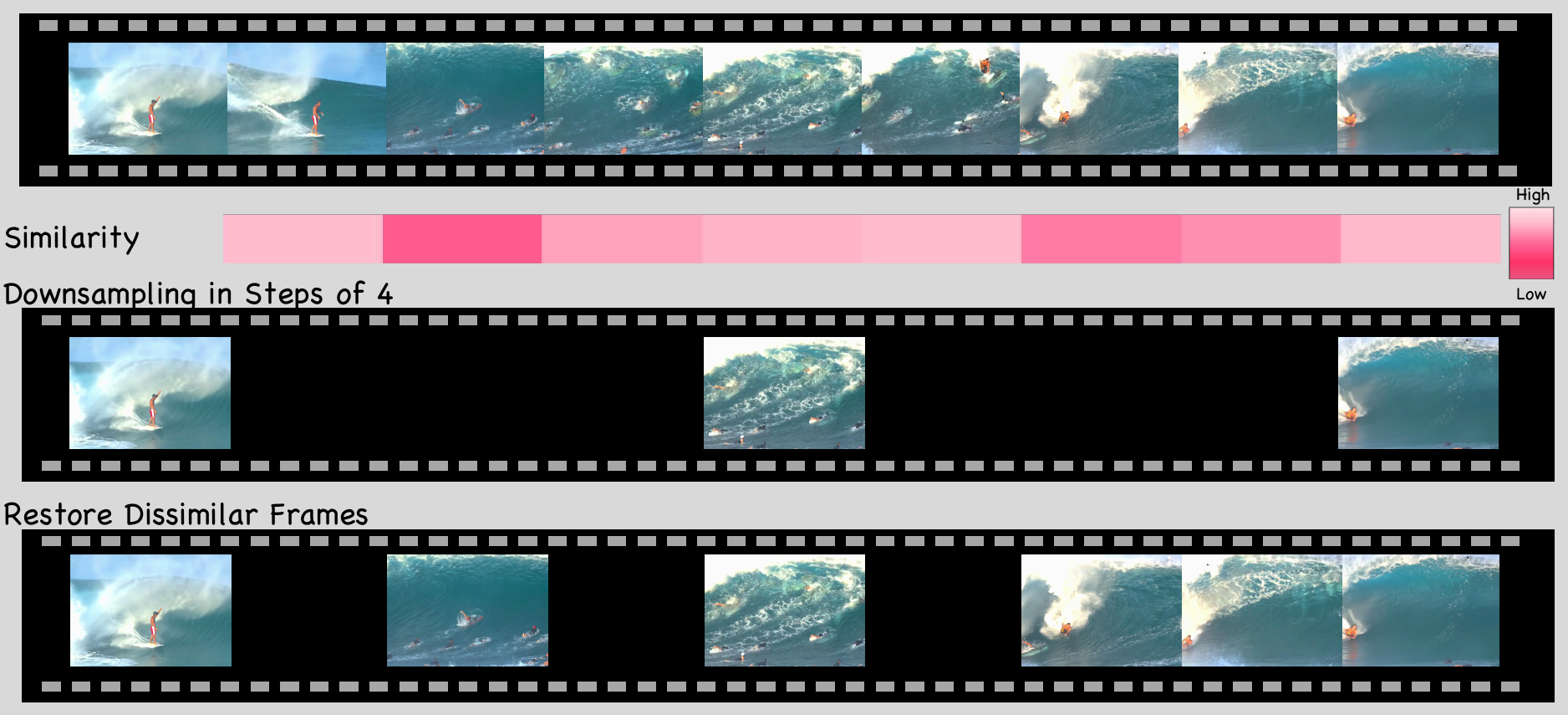}
    \vspace{-1.3em}
    \caption{\textbf{Visualization of Similarity-Based Downsampling.} Keyframes are preserved by restoring frames with low similarity after uniform downsampling. The similarity score for each frame is computed to its preceding frame.}
    \Description{A visual example of the dynamic downsampling process. Uniform downsampling first removes frames, and then low-similarity frames are restored as keyframes so that important temporal changes are preserved.}
    \vspace{-0.8em}
    \label{dynamic_downsampling}
\end{figure}

\noindent \textbf{Dynamic Downsampling Visualization.}
Figure~\ref{dynamic_downsampling} shows the qualitative effect of our frame selection strategy.
Stepwise downsampling removes redundant frames while preserving the basic structure of the video.
Restoring dissimilar frames recovers frames with low similarity and helps preserve critical temporal information.
Notably, the restored frames often coincide with scene transitions, object interactions, or motion onsets, indicating that the similarity-based criterion is effective at recovering visually informative moments that uniform sampling alone would miss.
The visualization shows that our approach balances frame reduction with information retention, which remains important for downstream video analysis and reasoning tasks.

\section{Conclusion}
This paper presents PCA, a training-free and plug-and-play framework for dynamically compressing video frames in VLLMs, aiming to enhance both efficiency and performance. 
To our knowledge, few works have explored dynamic frame pruning specifically for efficient VLLMs.
To fill this gap, PCA operates in two stages.
It first prunes redundant frames and then enhances each retained frame by aggregating contextual information from temporal neighbors, making the frames persistence-aware and temporally enriched for robust downstream reasoning.
Extensive experiments show that PCA consistently outperforms state-of-the-art baselines in both accuracy and efficiency, achieving a speedup of 1.8$\times$ to 2.5$\times$.
In future work, we will explore persistence-aware motion enhancement during training and study how it interacts with learned temporal representations.

\begin{acks}
\emergencystretch=1em
This work was supported by the National Natural Science Foundation of China (Grant No.\ 62576076), the CCF-Tencent Rhino-Bird Open Research Fund, the Guangdong Research Team for Communication and Sensing Integrated with Intelligent Computing (Project No.\ 2024KCXTD047), the Guangdong Basic and Applied Basic Research Foundation (Grant Nos.\ 2023A1515140037 and 2025B1515120017), and the Guangdong Provincial Key Laboratory (Grant No.\ 2023B1212060076). The computational resources are supported by SongShan Lake HPC Center (SSL-HPC) in Great Bay University.
\end{acks}

{
  \small
  \bibliographystyle{ACM-Reference-Format}
  \bibliography{main}
}

\end{document}